%% file: LatentIndex_arxiv0921.tex
\documentclass{article}
\usepackage{iclr2027_conference,times}

\input{math_commands.tex}

\usepackage{hyperref}
\usepackage{url}

\usepackage[utf8]{inputenc}
\usepackage[T1]{fontenc}
\usepackage{booktabs}
\usepackage{amsfonts}
\usepackage{nicefrac}
\usepackage[expansion=false]{microtype}
\usepackage{xcolor}
\usepackage{amsmath}
\usepackage{amssymb}
\usepackage{mathtools}
\usepackage{algorithm}
\usepackage{algorithmic}
\usepackage{graphicx}
\usepackage{multirow}
\usepackage{subcaption}
\usepackage{wrapfig}
\usepackage{tikz}
\usepackage{pgfplots}
\pgfplotsset{compat=1.18}
\usepackage{tabularx}
\usepackage{array}

\newcommand{\tbd}[1][]{\textcolor{blue}{[TBD\ifx&#1&\else: #1\fi]}}
\newcommand{\paperfigure}[2][\linewidth]{%
  \IfFileExists{#2}{\includegraphics[width=#1]{#2}}{%
    \fbox{\parbox[c][2.6cm][c]{0.92\linewidth}{\centering
      Figure asset required:\\ \texttt{\detokenize{#2}}}}}}

\title{LatentIndex: Cross-Layer Sharing with Layer-Specific Selection for Sparse Attention}
\author{%
\textbf{Zhaohui Wang}\textsuperscript{1,}\thanks{Equal contribution.},
\textbf{Zhixin Pan}\textsuperscript{2,3,}\footnotemark[1],
\textbf{Fanxu Meng}\textsuperscript{2,}\footnotemark[1]\thanks{Project leader. Email: \href{mailto:mengfanxu@xiaohongshu.com}{\texttt{mengfanxu@xiaohongshu.com}}.},
\textbf{Muhan Zhang}\textsuperscript{1,}\thanks{Corresponding author: \href{mailto:muhan@pku.edu.cn}{\texttt{muhan@pku.edu.cn}}.}\\[0.5em]
\textsuperscript{1}Institute for Artificial Intelligence, Peking University\\
\textsuperscript{2}Dots Studio, Xiaohongshu Inc.\\
\textsuperscript{3}Beijing Institute of Technology
}
\hypersetup{pdfauthor={Zhaohui Wang, Zhixin Pan, Fanxu Meng, Muhan Zhang}}

\iclrfinalcopy 
\makeatletter
\renewcommand{\@maketitle}{\vbox{\hsize\textwidth\centering
  \def\@makefnmark{\hbox{\@textsuperscript{\normalfont\@thefnmark}}}%
  {\LARGE\scshape\@title\par}
  \vskip 0.15in
  {\begin{tabular}[t]{c}\@author\end{tabular}\par}
  \vskip 0.6em
  {\normalfont\small\url{https://github.com/MuLabPKU/TransArch}\par}
  \vskip 0.3in minus 0.1in
}}
\makeatother
\begin{document}

\maketitle
\lhead{} 

\begin{abstract}

Sparse attention reduces core-attention computation, but its indexers still incur repeated selection work and per-layer key-cache storage. Reusing selected indices across layers reduces this overhead but constrains multiple layers to the same token set. We introduce LatentIndex, which extends the latent-sharing principle of Multi-head Latent Attention across indexer layers. Each layer group constructs a shared latent cache from its first layer’s hidden states, while layer-specific scoring enables independent token selection. Absorbing key decoders into queries enables direct scoring of the shared cache without reconstructing historical per-layer keys. We develop training-free calibration and investigate a training-aware instantiation of this principle. To balance quality and computation, a hierarchical selection (HS) variant lets followers independently refine a shared candidate set proposed by the anchor. With four-layer sharing, LatentIndex reduces logical indexer-cache storage by 61.1\% on DeepSeek-V3.2. Across DeepSeek-V3.2 and GLM-5, training-free LatentIndex improves head-wise attention-mass recall over IndexCache by up to 3.28 percentage points while maintaining RULER and LongBench performance close to native DSA. HS further achieves 2.30–2.72× decode indexer speedups over DSA across 8K–128K contexts, retaining most of LatentIndex’s recall. LatentIndex offers a new perspective on cross-layer indexing: sharing continuous representations rather than discrete selections enables efficient reuse while preserving layer-specific token selection.
\end{abstract}

\section{Introduction}
\label{sec:intro}
Long-context processing is increasingly important for large language models. Web-based agents and software-engineering workflows require models to integrate information across multiple interaction steps and diverse sources~\citep{nakano2021webgpt,yang2024swebench}, while extended reasoning further increases context lengths. These workloads make attention efficiency critical to inference speed and serving capacity. Dense attention requires $\mathcal{O}(L^2)$ computation over a sequence of length $L$; during autoregressive decoding, each new query also attends to a growing history, increasing computation and KV-cache traffic.

Sparse attention reduces this cost by restricting the tokens or blocks that each query attends to~\citep{child2019sparse,beltagy2020longformer,kitaev2020reformer,yuan2025nsa,lu2025moba}. DeepSeek Sparse Attention (DSA)~\citep{deepseek2025v32} implements fine-grained token selection through a lightning indexer, which scores historical tokens and selects the top-$k$ positions for core attention. This reduces core-attention complexity to $\mathcal{O}(Lk)$ over the sequence. However, the indexer scans the full prefix at each layer, retaining quadratic prefill complexity and maintaining a separate key cache. Sparse core attention therefore leaves two sources of indexing overhead: repeated selection computation and per-layer key storage.

Cross-layer redundancy offers an opportunity to reuse indexing information. IndexCache~\citep{bai2026indexcache} reports substantial overlap between nearby layers' top-$k$ selections and exploits it by retaining indexers in a subset of Full layers while letting Shared layers reuse their selected indices. This reduces indexing computation but constrains multiple layers to the same token set, preventing followers from adapting their selections to their own queries and relevance distributions. This motivates our central question: \textit{Can we exploit cross-layer redundancy while retaining layer-specific token selection?}

Multi-head Latent Attention (MLA)~\citep{deepseek2024v2} offers a useful design principle: a compact latent representation can be shared across attention heads while supporting head-specific attention. We extend this principle to indexer layers with \textbf{LatentIndex}. Each group of consecutive layers constructs one shared latent cache from the hidden states of its first layer, termed the anchor. Layer-specific decoders define effective indexer keys for all group members, including the anchor, while each layer retains its own scoring and top-$k$ selection. At inference time, the decoders are absorbed into queries, enabling direct scoring of the shared cache without reconstructing historical per-layer keys. This design replaces separate indexer key caches with a shared representation while leaving the main-attention KV caches unchanged.

To further accelerate indexing while preserving layer-specific decisions, we introduce a \emph{hierarchical selection} (HS) variant. The anchor scans the full prefix to form a broad candidate set, and each follower independently scores these candidates and selects its own top-$k$ tokens. HS thus balances computational savings with layer-specific selection quality.

We instantiate LatentIndex in both training-free and training-aware settings. Training-free calibration uses offline ridge reduced-rank regression to jointly fit shared projections and layer-specific decoders, yielding a shared representation that serves the entire group. All fitted parameters remain fixed during inference. Training-aware adaptation begins with a frozen backbone, training each layer's indexer to match its dense-attention distribution. It then updates the backbone through sparse language modeling while continuing attention-based supervision of the indexers, without changing the main-attention architecture.

We evaluate LatentIndex in both settings. 
On DeepSeek-V3.2 and GLM-5, training-free LatentIndex improves head-wise attention-mass recall over IndexCache at every evaluated length from 4K to 128K, by up to 3.28 percentage points. Its aggregate RULER and LongBench scores exceed IndexCache's on both models and remain close to native DSA's. With four-layer sharing on DeepSeek-V3.2, LatentIndex reduces logical indexer-cache storage by 61.1\%. HS retains most of its recall while achieving $2.30$--$2.72\times$ decode indexer speedups over DSA across 8K--128K, measured across all 61 indexers. Compared with IndexCache, HS improves average recall, RULER, and LongBench scores through additional layer-specific candidate scoring. Training experiments on DeepSeek-V2-Lite further show higher warmup recall and generally lower sparse-training language-modeling loss than IndexCache.

Our contributions are threefold:
\begin{itemize}

\item We introduce LatentIndex, bringing MLA's latent-sharing principle to sparse-attention indexers across layers. 
Shared latent caching reduces indexer-cache storage, while query-side decoder absorption enables direct scoring with layer-specific token selection.

\item We develop training-free calibration and investigate training-aware adaptation. We further introduce hierarchical selection to reduce repeated full-prefix scoring through shared candidates and layer-specific refinement.

\item We demonstrate improved selection quality over index reuse, downstream performance close to native DSA, and a quality--latency trade-off through hierarchical selection. Training experiments further show that shared latent indexers can be learned while maintaining language-model loss close to native DSA.

\end{itemize}

\section{Related Work}
\label{sec:related}

\paragraph{Sparse Attention.}
Sparse attention reduces attention cost by restricting the positions accessed by each query through fixed patterns, local/global connectivity, or content-dependent selection~\citep{child2019sparse,beltagy2020longformer,kitaev2020reformer}.
Recent methods include Native Sparse Attention (NSA), MoBA, and MiniMax Sparse Attention, which combine selective access with efficient attention execution~\citep{yuan2025nsa,lu2025moba,lai2026msa}.
DeepSeek Sparse Attention (DSA)~\citep{deepseek2025v32} uses a lightning indexer to select tokens for the main attention computation. Although this reduces the main attention cost, the indexer still scores the full causal prefix for each query, retaining quadratic scoring complexity during prefill.

\paragraph{Efficient Indexing for Sparse Attention.}
Existing approaches improve indexing efficiency through cross-layer reuse, restricted search, and reduced representation or head costs.
IndexCache~\citep{bai2026indexcache} lets subsequent layers reuse an anchor layer's top-$k$ indices, while DeepSeek-V4.1-Flash~\citep{deepseekai2026deepseekv41flash} includes a Reuse mode for the available main KV cache and its latest selection.
SparDA~\citep{fu2026sparda} predicts the KV blocks needed by the next layer, enabling prefetching to overlap with current-layer execution.
HISA~\citep{xu2026hisa} selects candidate blocks before token-level refinement, and PIVOT~\citep{liu2026pivot} amortizes full-prefix scoring across nearby queries through a proxy query.
Other designs select over compressed representations~\citep{xu2026deepseek}, dynamically activate indexer heads~\citep{zhou2026misa}, or use fewer query heads~\citep{qiu2026design}.
LatentIndex explores cross-layer representation sharing as a basis for efficient indexing while preserving layer-specific token selection.

\paragraph{Representation Sharing and Compression.}

Sharing continuous representations offers another route to reducing attention overhead.
CLA~\citep{brandon2024cla} and MLKV~\citep{zuhri2025mlkv} share key/value heads across layers, while YOCO~\citep{sun2024yoco} constructs global KV representations for reuse across its cross-decoder layers.
KVSharer~\citep{yang2024kvsharer} explores post-hoc layer-wise cache sharing.
Beyond direct reuse, xKV~\citep{chang2025xkv} jointly factorizes grouped-layer KV caches into a shared low-rank subspace, and CommonKV~\citep{wang2025commonkv} factorizes adjacent-layer projection weights to facilitate latent-cache merging.
Dimensional compression also supports efficient attention and retrieval: MLA~\citep{deepseek2024v2} combines latent KV storage with query-side absorption of content-key projections, while Loki~\citep{singhania2024loki} exploits low-dimensional key structure for token selection.
LatentIndex extends representation sharing from main-attention KV representations to the key representations maintained across indexer layers.

\section{Method: LatentIndex}
\label{sec:method}

\begin{figure}[h]
\begin{center}
\paperfigure[\linewidth]{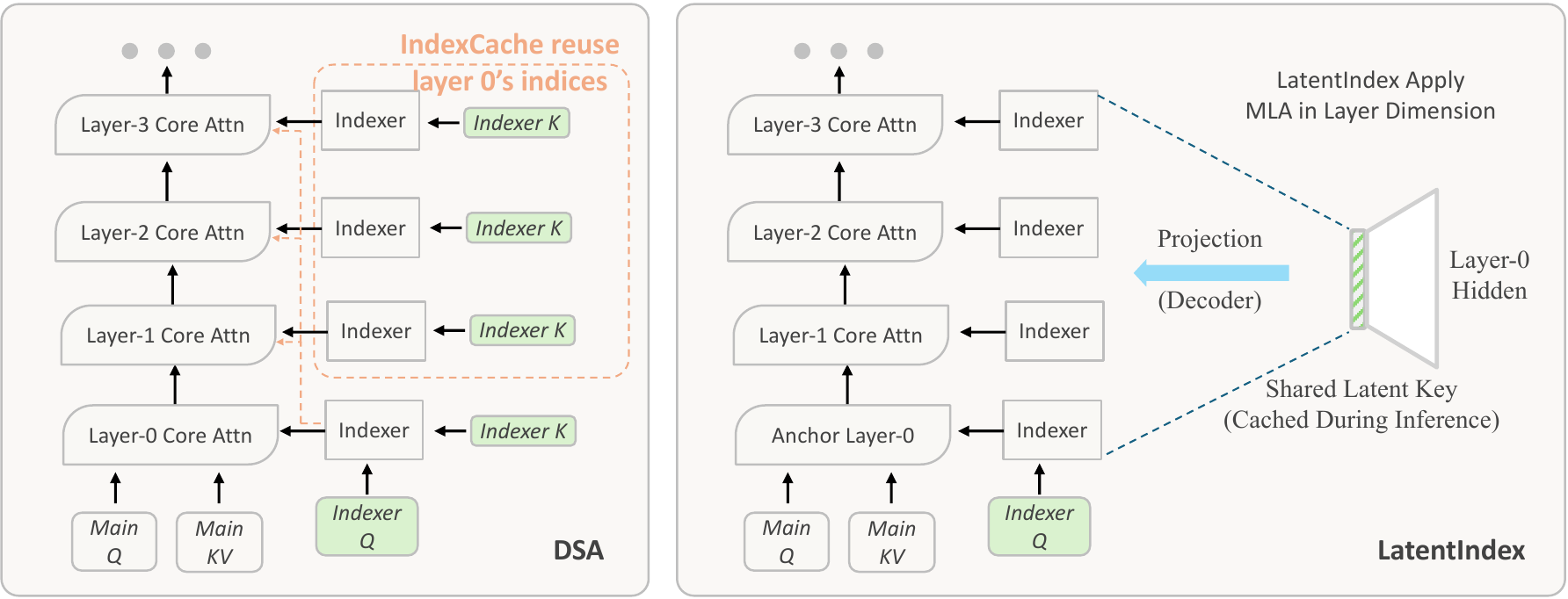}

\end{center}
\caption{
DSA uses per-layer indexer key caches and token selections; IndexCache reuses the anchor's selections (orange dashed paths). LatentIndex shares one latent cache derived from anchor hidden states, with layer-specific decoders defining layer-specific keys. At inference, decoders are absorbed into queries for direct cache scoring and independent token selection, without reconstructing historical keys. Queries are layer-specific but are illustrated only once for visual clarity.
}
\label{fig:architecture}
\end{figure}

LatentIndex (Figure~\ref{fig:architecture}) extends MLA's latent-sharing principle to the layer dimension of sparse-attention indexers, sharing key representations while retaining layer-specific token selection. 
Each group stores a shared latent cache, while layer-specific decoders define the effective keys used by each layer. At inference, these decoders are absorbed into queries, allowing each layer to score the shared cache directly without reconstructing historical keys.
We first describe the shared representation and its scoring procedure, then explain how to obtain the parameters through offline calibration or training and how hierarchical selection reduces repeated full-prefix scoring.

\subsection{Shared Key Representations with Layer-Specific Indexers}
\label{sec:ski_framework}

\paragraph{Notation.}

We partition indexer layers into consecutive groups $\{\mathcal G_g\}$ of at most $G$ layers, with the first layer $a_g=\min\mathcal G_g$ serving as the anchor. Let $\mathbf x_t^{(\ell)}$ denote the hidden state of token $t$ at layer $\ell$ after input RMSNorm, represented as a column vector. The native indexer key and shared latent have dimensions $d_I=d_R+d_N$ and $d_C=r_R+r_N$, respectively, where the subscripts $R$ and $N$ indicate RoPE and NoPE components. Each indexer uses $H_I$ query heads.

\paragraph{Shared Latent Cache.}
To replace separate per-layer key caches, each group constructs one shared representation per token from the anchor hidden state. This representation has a RoPE branch that encodes position and a NoPE branch. For token position $t$, we compute
\begin{equation}
\mathbf{z}^{(g),N}_t
=
\mathbf{P}_{g,N}^{\top}\mathbf{x}^{(a_g)}_t,
\qquad
\mathbf{z}^{(g),R}_t
=
\mathbf{Q}_t\mathbf{P}_{g,R}^{\top}\mathbf{x}^{(a_g)}_t,
\label{eq:ski_latent}
\end{equation}
where  
$\mathbf{P}_{g,N}\in\mathbb{R}^{d_{\mathrm{model}}\times r_N}$ and $\mathbf{P}_{g,R}\in\mathbb{R}^{d_{\mathrm{model}}\times r_R}$ are group-specific projections, and $\mathbf{Q}_t\in\mathbb{R}^{r_R\times r_R}$ applies rotary position encoding in the latent space. The two branches are concatenated and cached once per group:
\begin{equation}
\mathbf{c}^{(g)}_t
=
\begin{bmatrix}
\mathbf{z}^{(g),R}_t\\
\mathbf{z}^{(g),N}_t
\end{bmatrix}
\in\mathbb{R}^{d_C}.
\label{eq:ski_cache}
\end{equation}
The two branches form a single cache used by every group member. Each layer interprets this shared representation through its own decoder, rather than directly reusing the anchor's native key.

\paragraph{Layer-specific Key Reconstruction.}
To let a shared cache support different relevance scores across layers, each layer uses its own decoder to map the latent representation to effective indexer keys.

For each $\ell\in\mathcal G_g$, layer-specific decoders define the effective key components:
\begin{equation}
\widehat{\mathbf{k}}^{(\ell),R}_t
=
\mathbf{D}^{(\ell),R}\mathbf{z}^{(g),R}_t,
\qquad
\widehat{\mathbf{k}}^{(\ell),N}_t
=
\mathbf{D}^{(\ell),N}\mathbf{z}^{(g),N}_t
+
\mathbf{b}^{(\ell)},
\label{eq:latentindex_key_components}
\end{equation}
where $\mathbf{D}^{(\ell),R}\in\mathbb{R}^{d_R\times r_R}$,
$\mathbf{D}^{(\ell),N}\in\mathbb{R}^{d_N\times r_N}$, and
$\mathbf{b}^{(\ell)}\in\mathbb{R}^{d_N}$.
Because the rotary latent is already position-encoded when cached, its decoder must preserve the native rotary structure. We therefore require decoding the rotated latent to give the same result as rotating the decoded key:
\begin{equation}
\mathbf R_t\mathbf D^{(\ell),R}
=
\mathbf D^{(\ell),R}\mathbf Q_t,
\label{eq:latentindex_rope_commutation}
\end{equation}
for all $t$, where $\mathbf R_t\in\mathbb R^{d_R\times d_R}$ is the native rotary map.
This condition makes decoding before native rotation equivalent to decoding after latent rotation.
We enforce it through complex linear combinations within each rotary frequency, without cross-frequency mixing or a rotary bias
(Appendix~\ref{app:rope_decoder}).

Combining both components gives
\begin{equation}
\widehat{\mathbf{k}}^{(\ell)}_t
=
\mathbf{D}^{(\ell)}\mathbf{c}^{(g)}_t
+
\mathbf{b}^{(\ell)}_{\mathrm{ext}},
\qquad
\mathbf{D}^{(\ell)}
=
\begin{bmatrix}
\mathbf{D}^{(\ell),R} & \mathbf{0}\\
\mathbf{0} & \mathbf{D}^{(\ell),N}
\end{bmatrix},
\qquad
\mathbf{b}^{(\ell)}_{\mathrm{ext}}
=
\begin{bmatrix}
\mathbf{0}\\
\mathbf{b}^{(\ell)}
\end{bmatrix},
\label{eq:latentindex_decoder}
\end{equation}
with $\mathbf D^{(\ell)}\in\mathbb R^{d_I\times d_C}$.
These effective keys define layer-specific scoring. 
Note that during actual inference, the decoders are absorbed into the queries, avoiding explicit key reconstruction.

\paragraph{Query-side Decoder Absorption.}
Explicitly reconstructing historical keys would introduce additional computation and memory overhead, but indexing only requires their dot products with the current queries. 
As in MLA, we absorb each layer's key decoder into its queries to score the shared cache directly.
Let $\mathbf q^{(\ell)}_{s,h}\in\mathbb R^{d_I}$ denote the native indexer query at position $s$ and head $h$, with native RoPE already applied to its rotary component.
Using the decoder and bias defined above, we compute
\begin{equation}
\widetilde{\mathbf q}^{(\ell)}_{s,h}
=
\left(\mathbf D^{(\ell)}\right)^\top
\mathbf q^{(\ell)}_{s,h}
\in\mathbb R^{d_C},
\qquad
\beta^{(\ell)}_{s,h}
=
\left(\mathbf q^{(\ell),N}_{s,h}\right)^\top
\mathbf b^{(\ell)},
\label{eq:ski_query}
\end{equation}
where $\mathbf q^{(\ell),N}_{s,h}$ is the query's NoPE component.
For any historical position $t\le s$, the query--key dot product satisfies
\begin{equation}
\left(\mathbf q^{(\ell)}_{s,h}\right)^\top
\widehat{\mathbf k}^{(\ell)}_t
=
\left(\widetilde{\mathbf q}^{(\ell)}_{s,h}\right)^\top
\mathbf c^{(g)}_t
+
\beta^{(\ell)}_{s,h}.
\label{eq:ski_absorption}
\end{equation}
Thus, each decoder is applied once per query head, and the resulting query scores the shared cache without reconstructing historical keys.
The bias contribution $\beta^{(\ell)}_{s,h}$ is shared across historical positions but must be added before the indexer's ReLU.
This reformulation is exact for the effective keys in exact arithmetic.

\paragraph{Layer-specific Scoring and Selection.}
Each layer uses its own head weights $w^{(\ell)}_{s,h}$ and indexer scaling to score the shared cache over valid causal positions $\mathcal V_s$:

\begin{equation}
u^{(\ell)}_{s,t}
=
\sum_{h=1}^{H_I}
w^{(\ell)}_{s,h}
\operatorname{ReLU}\!\left(
\left(\widetilde{\mathbf q}^{(\ell)}_{s,h}\right)^\top
\mathbf c^{(g)}_t
+
\beta^{(\ell)}_{s,h}
\right),
\qquad t\in\mathcal V_s.
\label{eq:latentindex_score}
\end{equation}
Each layer independently selects
\begin{equation}
\mathcal S^{(\ell)}_s
=
\operatorname{TopK}_{t\in\mathcal V_s}
\left(u^{(\ell)}_{s,t},\min(k,|\mathcal V_s|)\right),
\label{eq:latentindex_topk}
\end{equation}
which its main attention uses with its own KV cache.
Thus, layers share the latent cache while retaining layer-specific scores and token selections.

\subsection{Training-Free LatentIndex}
\label{sec:latentindex_training_free}

\paragraph{Offline Calibration.}
We instantiate LatentIndex in a frozen pretrained model through offline calibration of the shared projections and layer-specific decoders.
For each group, we collect the anchor hidden states and the native indexer keys of all group members at matching token positions. We treat the group members' keys as joint targets and use ridge reduced-rank regression~\cite{izenman1975reduced} to jointly fit a shared projection from the anchor hidden states and a separate decoder for each layer. The objective minimizes key-reconstruction error under the prescribed latent-rank constraints, with ridge regularization on the resulting mappings.

We fit the NoPE and RoPE branches separately. For NoPE, we concatenate the group members' target keys and fit a mapping of rank at most $r_N$ with an output bias. For RoPE, we fit a bias-free complex reduced-rank mapping independently at each native frequency. Same-frequency decoding preserves compatibility with positional rotation. Both branches admit closed-form solutions, requiring no gradient-based optimization or updates to pretrained weights.
The fitted projections generate the shared latent cache, and the fitted decoders define the query transformations described above. All fitted parameters remain fixed during inference. Appendix~\ref{app:latentindex_offline_fit} provides more details.

\paragraph{Hierarchical Selection.}
\label{sec:hierarchical_selection}

To further accelerate indexing while preserving layer-specific token selection, we restrict follower layers to a shared candidate set, leaving the latent representation and query-side decoder absorption unchanged. 
For each query position $s$, the anchor $a_g$ scores all valid causal positions $\mathcal V_s$ using Eq.~\eqref{eq:latentindex_score} and forms a candidate set $\mathcal C_s^{(g)}$ from the top-$B$ positions, where $B\ge k$. The anchor selects its own top-$k$ positions under the same scores. Each follower $\ell\in\mathcal G_g\setminus\{a_g\}$ independently rescores these candidates using its layer-specific scorer and selects its own top-$k$ set $\mathcal S_s^{(\ell)}$. Both budgets are capped by $|\mathcal V_s|$, and each layer uses its selected positions with its own main-attention KV cache. Thus, each follower scores at most $B$ positions per query while retaining layer-specific selection within the shared candidate set.

\subsection{Training-Aware LatentIndex}
\label{sec:latentindex_training_aware}
We also study training LatentIndex on DeepSeek-V2-Lite.
Training proceeds in two stages: dense-attention warmup learns the indexers with a frozen backbone, followed by sparse training that adapts the backbone and indexers using their selected routes.

\paragraph{Training instantiation.}
In this instantiation, the anchor hidden state supplies both the shared key representation and the inputs to all layer-specific query and head-weight projections.
Each layer uses its own NoPE decoder, query projection, and head-weight projection to compute relevance scores and independently select tokens. On DeepSeek-V2-Lite, we partition the 27 layers into nine consecutive groups of three, with top-2048 selection. For this training instantiation, we apply LayerNorm to the projected NoPE latent in Eq.~\eqref{eq:ski_latent}, use bias-free NoPE decoders, and fix the rotary decoder to the identity.
Because all routing inputs are available at the anchor, the group's routes can be computed at its entrance and then consumed by their respective layers during sparse execution.

\paragraph{Dense-attention warmup.}
We first freeze the pretrained backbone and train the indexers to match each layer's dense attention.
For each query, the teacher distribution averages the core-attention probabilities across heads, while the indexer distribution applies softmax to its scores over the full causal prefix.
Training minimizes the teacher-to-indexer KL divergence over valid prediction positions. Each layer's dense-attention distribution supervises its LatentIndex scoring distribution. In IndexCache, all three layer-specific teacher distributions supervise the anchor indexer.
The method-specific loss normalization is detailed in Appendix~\ref{app:training_implementation}.
This stage learns token-selection scores from dense attention while leaving the backbone and its dense-attention computation unchanged.

\paragraph{Sparse training.}
Starting from each method's warmup checkpoint, we activate its top-2048 selections and train the backbone with next-token language modeling, retaining the main-attention architecture.
The indexers continue to receive KL supervision, now using each layer's sparse-attention probabilities, with both distributions normalized over the selected positions.
Teacher targets and indexer inputs are detached from the backbone, and the top-$k$ operation does not pass language-modeling gradients to the selector.
Additional training details are provided in Appendix~\ref{app:training_details}.

\subsection{Storage and Computation}
\label{sec:ski_cost}
\paragraph{Indexer cache storage.}
For a group of $m$ layers, we store one $d_C$-dimensional latent instead of $m$ separate $d_I$-dimensional keys, giving a payload ratio of $d_C/(md_I)$. Including FP8 scales and the independent native layer, our DeepSeek-V3.2 configuration reduces logical indexer-cache storage by 61.1\%. The main-attention KV cache remains unchanged.

\paragraph{Fixed parameters and computation.}
Projection and decoder parameters are independent of context length. We generate the shared latent once per group and absorb each layer's decoder into its queries. Without hierarchical selection, each layer independently scores the full causal history, requiring $O(H_I Ld_C)$ work per query and $O(H_I L^2d_C)$ over prefill. With candidate budget $B$, hierarchical selection reduces the number of scored positions per query in a group of $m$ layers from $mL$ to $L+(m-1)\min(B,L)$. Appendix~\ref{app:storage_details} details storage accounting and inference implementation.

\section{Experiments}
\label{sec:experiments}

\subsection{Experimental Setup}
\label{sec:experimental_setup}
We compare native DSA, IndexCache, and LatentIndex on DeepSeek-V3.2~\citep{deepseek2025v32} and GLM-5~\citep{glm5team2026glm5vibecodingagentic}.
Grouped methods use identical layer boundaries with $G=4$ unless otherwise stated, and all methods select up to 2048 tokens per query.
Training-free LatentIndex uses offline calibration and segmented FP8 inference, with calibration data disjoint from evaluation data.
We evaluate token-selection quality using attention-mass recall and downstream performance using RULER~\citep{hsieh2024ruler} and LongBench~\citep{bai2024longbench}.
RULER covers six context lengths from 4K to 128K, while LongBench reports six task categories.
Full configurations and evaluation protocols are provided in Appendix~\ref{app:experimental_details}.

\subsection{Training-Free Token-Selection Quality}
\label{sec:selection_quality}

\paragraph{Metric.}
We measure how well an indexer's selected tokens capture the importance assigned by core attention.
Attention-mass recall directly measures how well the selected tokens preserve the attention mass prioritized by core attention at a fixed token budget.
For a fixed query and layer, let $p_{h,i}$ denote core-attention head $h$'s dense attention probability over the full causal prefix, computed using the native attention scale.
Let $S$ be the indexer's selected set of $K=\min(2048,N)$ tokens for prefix length $N$, and let $T_h=\operatorname{TopK}_i(p_{h,i},K)$ be head $h$'s oracle selection.
Head-wise attention-mass recall is
\begin{equation}
R_{\mathrm{head}}
=
\frac{1}{H}\sum_{h=1}^{H}
\frac{\sum_{i\in S}p_{h,i}}
{\max\!\left(\sum_{i\in T_h}p_{h,i},\epsilon\right)},
\qquad \epsilon=10^{-8},
\end{equation}
where $H$ is the number of core-attention heads.
This metric averages the fraction of each head's oracle attention mass captured at the same token budget, giving all heads equal weight.
Sampling and aggregation details are in Appendix~\ref{app:experimental_details}.

\paragraph{Results.}
LatentIndex outperforms IndexCache at every evaluated length on both models, with gains increasing toward longer contexts (Table~\ref{tab:selection_recall_head}).
At 128K, it improves recall by 3.28 percentage points on DeepSeek-V3.2 and 1.57 points on GLM-5.
These results support sharing continuous representations while retaining layer-specific selection to preserve attention mass.

\begin{table}[t]
\centering
\caption{Head-wise attention-mass recall on a common native DSA trajectory for each model.}
\label{tab:selection_recall_head}
\vspace{0.5em}

\scalebox{0.78}{
\begin{tabular}{lcccccc|cccccc}
\toprule
\multirow{2}{*}{\textbf{Method}}
& \multicolumn{6}{c|}{\textbf{DeepSeek-V3.2}}
& \multicolumn{6}{c}{\textbf{GLM-5}} \\
\cmidrule(lr){2-7}
\cmidrule(lr){8-13}
& 4K & 8K & 16K & 32K & 64K & 128K
& 4K & 8K & 16K & 32K & 64K & 128K \\
\midrule

DSA
& \textbf{99.33}
& \textbf{97.40}
& \textbf{95.09}
& \textbf{93.48}
& \textbf{91.10}
& \textbf{90.00}
& \textbf{99.45}
& \textbf{98.33}
& \textbf{97.21}
& \textbf{96.37}
& \textbf{95.16}
& \textbf{94.30} \\

IndexCache
& 98.86
& 95.86
& 92.49
& 90.26
& 86.68
& 85.06
& 99.16
& 97.49
& 95.78
& 94.65
& 92.79
& 91.55 \\

LatentIndex
& \underline{99.21}
& \underline{96.97}
& \underline{94.34}
& \underline{92.40}
& \underline{89.61}
& \underline{88.34}
& \underline{99.32}
& \underline{97.96}
& \underline{96.65}
& \underline{95.67}
& \underline{94.22}
& \underline{93.12} \\

\midrule
$\Delta$ vs. IndexCache
& +0.35
& +1.11
& +1.85
& +2.14
& +2.93
& +3.28
& +0.16
& +0.47
& +0.87
& +1.02
& +1.43
& +1.57 \\

\bottomrule
\end{tabular}
}
\end{table}

\begin{table}[t]
\centering
\caption{RULER performance across context lengths. All grouped methods use $G=4$.}
\label{tab:ruler}
\small
\begin{tabular}{llccccccc}
\toprule

Model & Method
& 4K & 8K & 16K
& 32K & 64K & 128K
& Avg. \\
\midrule

\multirow{3}{*}{DeepSeek-V3.2}
& DSA
& \underline{97.21}
& 96.84
& \textbf{97.51}
& \textbf{97.46}
& \underline{95.89}
& \textbf{94.30}
& \textbf{96.54} \\

& IndexCache
& \textbf{97.28}
& \textbf{96.94}
& \textbf{97.51}
& 97.18
& \textbf{95.91}
& 93.59
& 96.40 \\

& LatentIndex
& 97.18
& \underline{96.92}
& \underline{97.35}
& \underline{97.42}
& 95.85
& \underline{94.22}
& \underline{96.49} \\

\midrule

\multirow{3}{*}{GLM-5}
& DSA
& \textbf{98.07}
& \textbf{97.57}
& \textbf{97.69}
& \textbf{97.41}
& \underline{97.40}
& \underline{96.74}
& \textbf{97.48} \\

& IndexCache
& 97.73
& \underline{97.42}
& 97.51
& 96.92
& 95.70
& 90.15
& 95.91 \\

& LatentIndex
& \underline{97.89}
& \textbf{97.57}
& \underline{97.64}
& \underline{97.36}
& \textbf{97.42}
& \textbf{96.81}
& \underline{97.45} \\

\bottomrule
\end{tabular}
\end{table}

\begin{table}[t]
\centering
\caption{LongBench performance across six task categories.}
\label{tab:longbench}
\small
\begin{tabular}{llccccccc}
\toprule

Model & Method
& SQA & MQA & Sum
& FS & Syn & Code
& Avg. \\
\midrule

\multirow{3}{*}{DeepSeek-V3.2}
& DSA
& \textbf{52.05}
& \textbf{53.79}
& \textbf{22.42}
& \textbf{61.87}
& \underline{68.17}
& \underline{73.12}
& \underline{55.24} \\

& IndexCache
& 51.37
& 53.25
& 22.24
& 59.78
& 67.91
& 73.11
& 54.61 \\

& LatentIndex
& \underline{52.00}
& \underline{53.74}
& \underline{22.27}
& \underline{61.06}
& \textbf{68.78}
& \textbf{74.18}
& \textbf{55.34} \\

\midrule

\multirow{3}{*}{GLM-5}
& DSA
& \underline{53.09}
& \textbf{56.02}
& \underline{24.40}
& 68.45
& \textbf{76.50}
& \underline{72.09}
& \textbf{58.42} \\

& IndexCache
& 53.01
& 55.67
& \textbf{24.44}
& \textbf{69.28}
& 73.33
& \textbf{72.11}
& 57.97 \\

& LatentIndex
& \textbf{53.31}
& \underline{55.71}
& 24.35
& \underline{68.58}
& \underline{76.33}
& 71.34
& \underline{58.27} \\

\bottomrule
\end{tabular}
\end{table}

\subsection{Training-Free Downstream Performance}
\label{sec:long_context}

We further evaluate each method end-to-end, where its token selections affect subsequent hidden states and generation.

\paragraph{RULER.}
LatentIndex closely preserves native DSA performance, with average scores only 0.05 and 0.03 points lower on DeepSeek-V3.2 and GLM-5, respectively (Table~\ref{tab:ruler}).
It improves over IndexCache in aggregate on both models.
On DeepSeek-V3.2 at 128K, it outperforms IndexCache by 0.63 points and remains 0.08 below DSA.
The largest gain over IndexCache occurs on GLM-5 at 128K, where LatentIndex reaches 96.81, compared with 90.15 for IndexCache and 96.74 for DSA.

\paragraph{LongBench.}
Table~\ref{tab:longbench} covers single-document QA (SQA), multi-document QA (MQA), summarization (Sum), few-shot learning (FS), synthetic retrieval (Syn), and code completion (Code).
LatentIndex improves the average score over IndexCache by 0.73 points on DeepSeek-V3.2 and 0.30 points on GLM-5, while remaining close to native DSA.
LatentIndex improves all six categories relative to IndexCache on DeepSeek-V3.2, with the largest gain in few-shot learning.
On GLM-5, it improves three of the six categories, with the largest gain in synthetic retrieval.

\subsection{Analysis of Cross-Layer Sharing}
\label{sec:ablation}

\begin{figure}[t]
\centering
\begin{subfigure}[t]{0.53\textwidth}
    \centering
    \includegraphics[width=\linewidth]{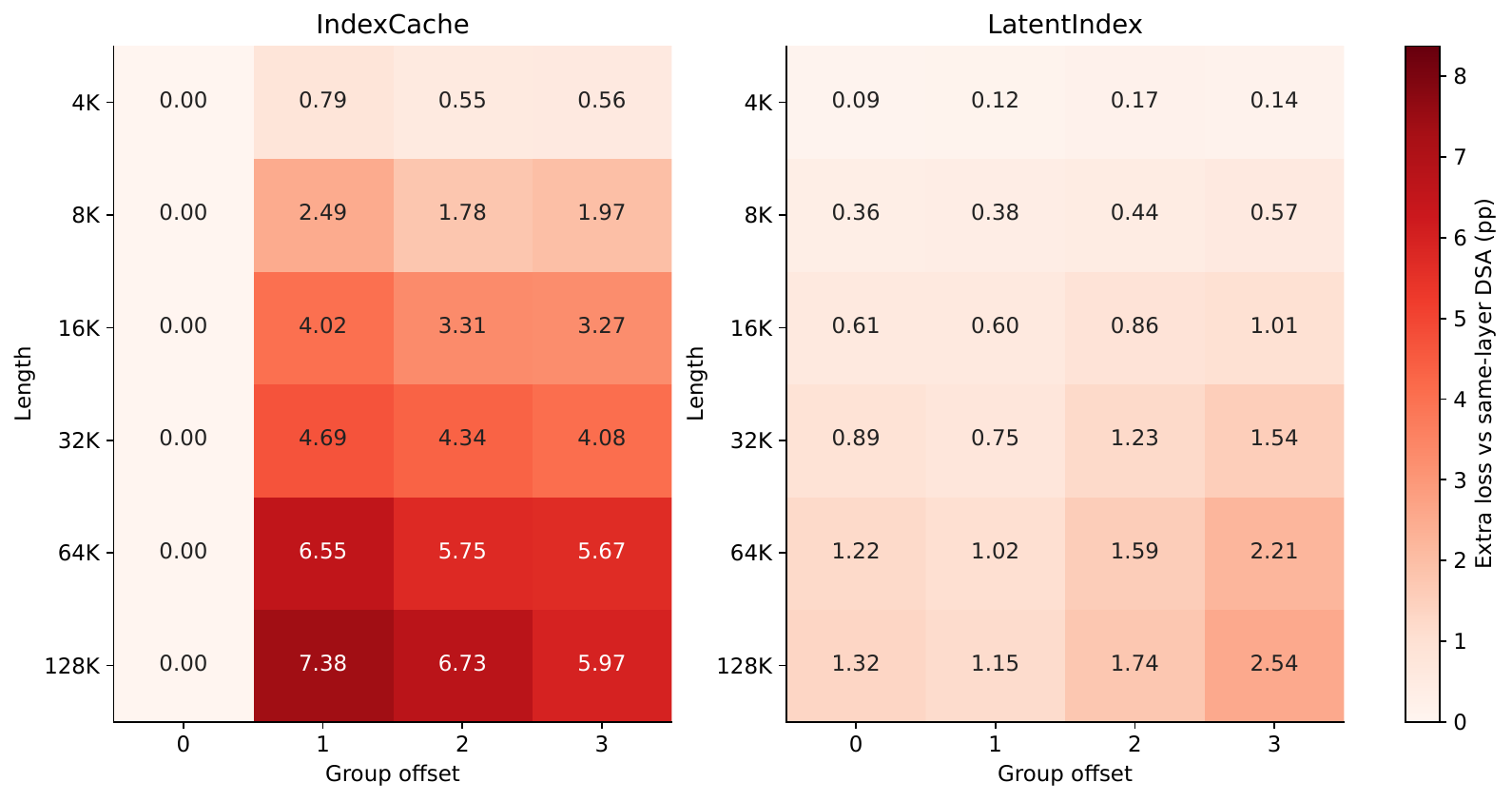}
    \caption{Recall loss by group offset}
    \label{fig:recall_loss}
\end{subfigure}
\hfill
\begin{subfigure}[t]{0.42\textwidth}
    \centering
    \includegraphics[width=\linewidth]{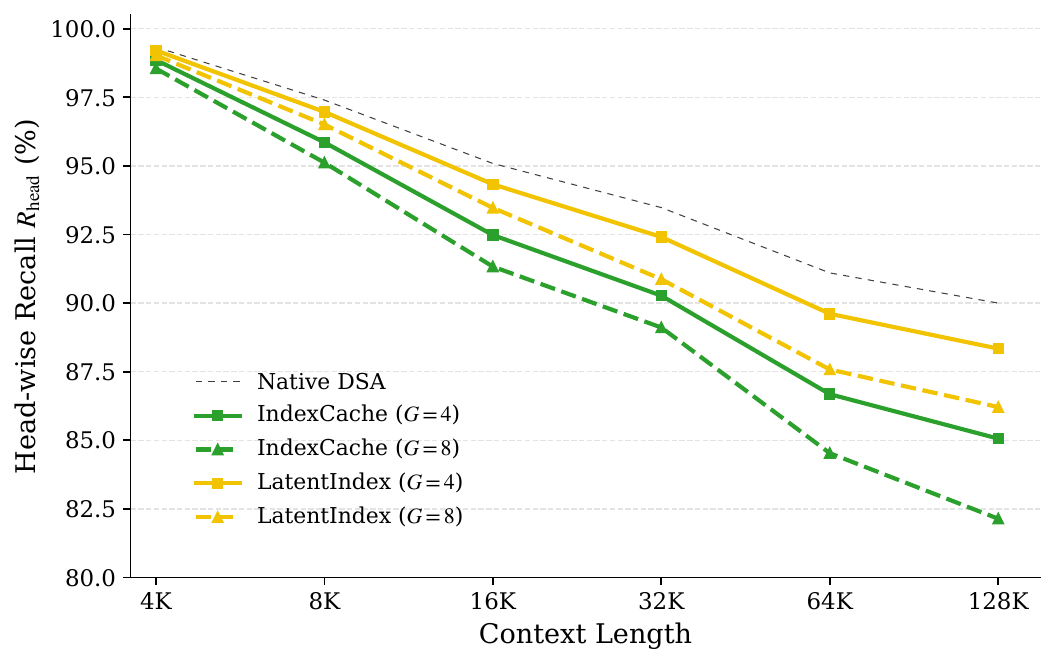}
    \caption{Effect of group size}
    \label{fig:ablation_groupsize}
\end{subfigure}
\caption{Cross-layer sharing on DeepSeek-V3.2: (a) additional $R_{\mathrm{head}}$ loss relative to native DSA by group offset ($G=4$); (b) $R_{\mathrm{head}}$ across context lengths for $G=4$ and $G=8$, with native DSA as a reference.
}
\label{fig:sharing_analysis}
\end{figure}

\paragraph{Selection quality across layers within a group.}
We measure additional recall loss relative to each layer's native DSA selection:
$
E_m^{(\ell)}
=
R_{\mathrm{head}}^{(\ell)}(S_{\mathrm{DSA}}^{(\ell)})
-
R_{\mathrm{head}}^{(\ell)}(S_m^{(\ell)}),
\qquad
m\in\{\mathrm{IndexCache},\mathrm{LatentIndex}\}.
\label{eq:ablation_extra_loss}
$
Within each four-layer group, offset 0 denotes the anchor and offsets 1--3 the followers.
IndexCache preserves the anchor's native selection but incurs an immediate recall loss when that selection is reused by followers (Figure~\ref{fig:recall_loss}).
This loss grows with context length at every follower offset, reaching 5.97--7.38 percentage points at 128K.
LatentIndex reduces recall loss at every follower offset and context length. At 128K, average follower loss decreases by 73\%, from 6.69 to 1.81 percentage points, while anchor loss is 1.32 points.
These results show that layer-specific selection mitigates the loss from cross-layer sharing, particularly at long contexts.

\paragraph{Effect of sharing group size.}
Figure~\ref{fig:ablation_groupsize} compares $G=4$ and $G=8$ with a fixed 192-dimensional latent.
Larger groups reduce recall for both methods, particularly at long contexts, while LatentIndex consistently outperforms IndexCache at either group size.
At 128K, LatentIndex with $G=8$ achieves 86.21 recall, exceeding IndexCache with $G=8$ by 4.07 percentage points and even IndexCache with $G=4$ by 1.15 points.
Increasing $G$ from 4 to 8 also reduces LatentIndex's shared caches from 15 to 8 and its logical indexer-cache storage from 3132 to 1732 bytes per token, supporting larger sharing groups while retaining higher selection quality than index reuse.

\subsection{Hierarchical Selection: Quality–Efficiency Trade-off}

We evaluate LatentIndex with hierarchical selection (HS) on DeepSeek-V3.2 using a fixed candidate budget of 8192.
Decode timing covers complete indexer forward calls across all 61 layers, including candidate construction and TopK but excluding main attention and other model components; timing details are in Appendix~\ref{app:hs_timing}.
HS  (Table~\ref{tab:hierarchical_selection_a}) achieves $R_{\mathrm{head}}=93.19$, compared with 93.48 for full-prefix LatentIndex, with RULER and LongBench scores of 96.58 and 54.81, respectively. It exceeds IndexCache on all three aggregate quality metrics and achieves $2.30$--$2.72\times$ decode indexer speedups over native DSA across 8K--128K. Relative to IndexCache, its additional indexer computation enables layer-specific refinement within shared candidates.

\begin{table}[t]
\centering
\caption{Hierarchical selection on DeepSeek-V3.2. Latency is aggregated over all 61 indexers in ms per decode token; parentheses show speedup over native DSA. Recall and RULER are averaged over six context lengths, and LongBench over six task categories.}
\label{tab:hierarchical_selection_a}
\small
\setlength{\tabcolsep}{3.4pt}
\resizebox{\textwidth}{!}{%
\begin{tabular}{lccc@{\hspace{8pt}}ccccc}
\toprule
& \multicolumn{3}{c}{Quality}
& \multicolumn{5}{c}{Decode indexer latency (ms) $\downarrow$} \\
\cmidrule(lr){2-4}
\cmidrule(lr){5-9}
Method
& $R_{\mathrm{head}}$ $\uparrow$
& RULER $\uparrow$
& LongBench $\uparrow$
& 8K & 16K & 32K & 64K & 128K \\
\midrule
Native DSA
& \textbf{94.40} & 96.54 & \textbf{55.24}
& 48.70 & 52.06 & 48.28 & 48.43 & 49.60 \\
IndexCache
& 91.54 & 96.40 & 54.61
& \textbf{14.31}
& \textbf{13.29}
& \textbf{14.35}
& \textbf{14.28}
& \textbf{13.40} \\
LatentIndex + HS
& 93.19 & \textbf{96.58} & 54.81
& 19.89\,(2.45$\times$)
& 19.13\,(2.72$\times$)
& 20.55\,(2.35$\times$)
& 21.10\,(2.30$\times$)
& 20.48\,(2.42$\times$) \\
\bottomrule
\end{tabular}%
}
\end{table}

\subsection{Training-Aware Evaluation}
\label{sec:training_aware}

\paragraph{Protocol.}
We compare DSA, IndexCache, and training-aware LatentIndex on DeepSeek-V2-Lite at 64K, using eight indexer heads, top-2048 selection, and $G=3$ for grouped methods.
All methods complete 400 steps of indexer-only dense warmup, followed by sparse training that updates both the backbone and indexers.
Sparse runs use matching samples at each step.
Training configurations, method-specific loss normalization, and diagnostics are detailed in Appendix~\ref{app:training_aware_details}.

\paragraph{Warmup selection quality.}

\begin{figure}[t]
\centering
\paperfigure[\linewidth]{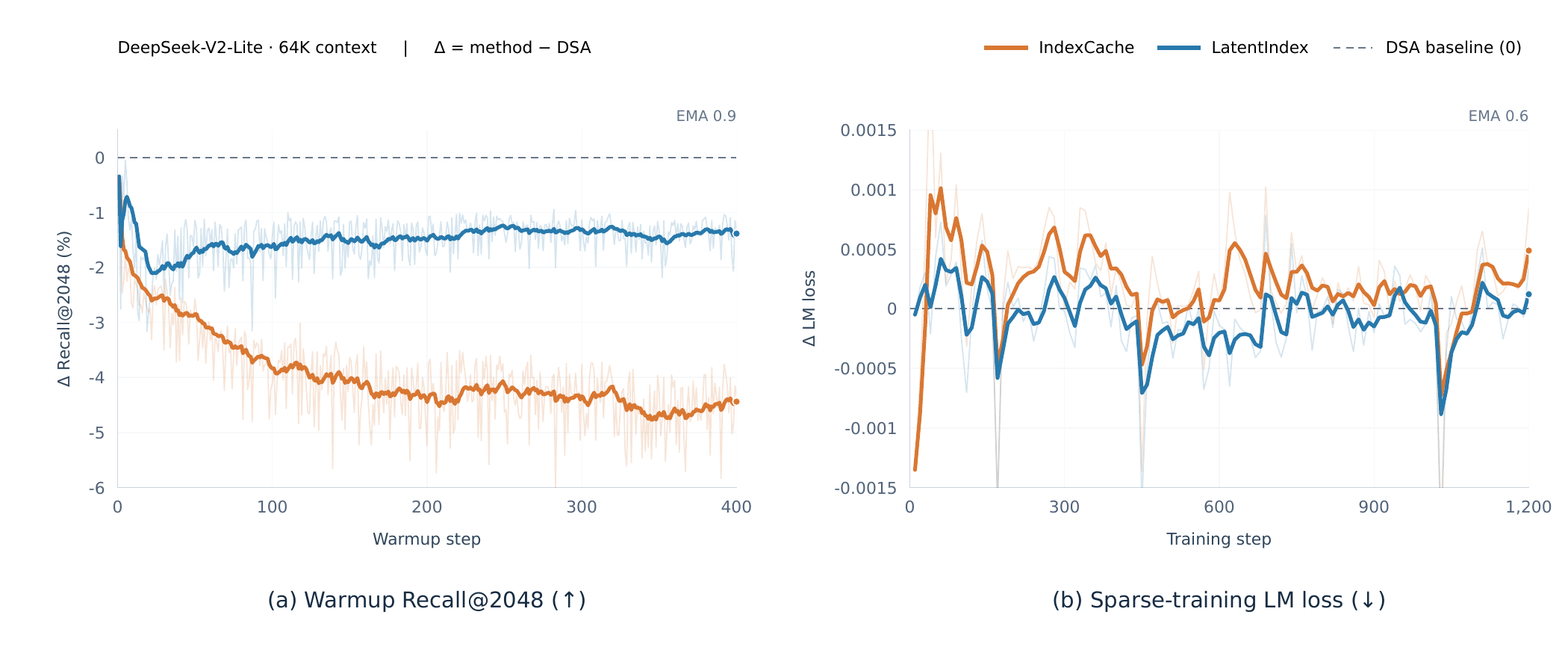}
\caption{Training dynamics on DeepSeek-V2-Lite at 64K: (a) warmup Recall@2048 and (b) sparse-training LM loss. Values are method minus DSA at matched steps; the dashed line marks zero difference. Faint and bold curves show raw and EMA-smoothed differences, respectively.}
\label{fig:training_aware}
\end{figure}

Figure~\ref{fig:training_aware} (a) reports matched-step Recall@2048 differences from DSA.
The metric measures captured attention mass relative to each head's top-2048 oracle, averaged over sampled queries and all 27 layers; diagnostic scaling and sampling are detailed in Appendix~\ref{app:training_diagnostics}.
After an initial widening, LatentIndex's gap to DSA narrows and stabilizes, whereas IndexCache's gap generally grows before leveling off.
In later warmup steps, LatentIndex remains approximately 1.3--1.5 percentage points below DSA, compared with 4.4--4.7 for IndexCache.
These results show that LatentIndex learns selections more closely aligned with core attention than index reuse.

\paragraph{Sparse-training behavior.}
Figure~\ref{fig:training_aware}(b) shows matched-step differences in training next-token cross-entropy.
LatentIndex generally maintains lower LM loss than IndexCache and fluctuates closely around the DSA baseline.
Together with its higher warmup recall, this indicates that shared latent indexing preserves selection quality while supporting sparse language-model training with loss close to DSA.

\section{Conclusion}
\label{sec:conclusion}
LatentIndex enables sparse-attention indexers to share continuous key representations across layers while retaining layer-specific token selection.
Training-free results on DeepSeek-V3.2 and GLM-5 show higher attention-mass recall than index reuse and downstream performance close to native DSA, with a 61.1\% reduction in logical indexer-cache storage on DeepSeek-V3.2.
The recall advantage is particularly pronounced for followers at long contexts and persists with larger sharing groups.
Hierarchical selection reduces repeated full-prefix scoring through shared candidates, achieving decode indexer speedups while retaining layer-specific refinement.
Training experiments provide complementary evidence that shared latent indexers can be learned while maintaining language-model loss close to DSA.
Together, these results support cross-layer representation sharing as a way to reduce indexer storage while preserving selection quality, with hierarchical selection providing additional computational savings.

\bibliography{iclr2027_conference}
\bibliographystyle{iclr2027_conference}


\appendix

\section{Offline Construction of LatentIndex}
\label{app:latentindex_offline_fit}

We construct training-free LatentIndex by jointly fitting a shared projection and layer-specific decoders to each group's native indexer keys, including those of the anchor. We fit the NoPE and RoPE branches separately using ridge reduced-rank regression.

\subsection{Calibration Data and Targets}
\label{app:calibration_data}

We freeze the pretrained backbone and collect features along its native DSA trajectory. At matching token positions, we record the anchor hidden states after input RMSNorm and each group member's native indexer keys after key normalization but before RoPE, Hadamard rotation, and quantization. We fit only key-reconstruction targets.

We draw calibration data from FineWeb-Edu, GitHub source code, OpenMathInstruct-2, SmolTalk, and PG19, covering general text, code, mathematics, dialogue, and long documents. Each domain contributes eight 8K and two 32K training requests, plus two 8K and one 32K validation requests. This yields 50 training and 15 validation requests, totaling 655,360 and 245,760 tokens under the DeepSeek-V3.2 tokenizer. General-text, mathematics, and dialogue requests may concatenate examples; each code or book request uses a single repository or book. We check source identities and text overlap across splits and keep calibration data separate from downstream evaluation. GLM-5 follows the same construction protocol.

All training prompt positions contribute to fitting. We use validation data to select regularization strengths and retain the training-fitted parameters without refitting on the combined splits. 

\subsection{RoPE-Compatible Decoders}
\label{app:rope_decoder}

\paragraph{Same-frequency decoding.}
For each native rotary frequency $f$, we construct $r_f$ complex latent sources $v_{t,f,j}$ from the anchor hidden state and reconstruct layer $\ell$'s pre-RoPE key coordinate as
\begin{equation}
\widehat{\overline{k}}_{t,f}^{(\ell),R}
=\sum_{j=1}^{r_f}\gamma_{f,j}^{(\ell)}v_{t,f,j}.
\label{eq:app_rotary_reconstruction}
\end{equation}
We apply the same native rotary phase to all sources associated with $f$. Complex linear decoding within each frequency, without cross-frequency mixing or an output bias, then satisfies
\begin{equation}
\mathbf R_t\mathbf D^{(\ell),R}
=\mathbf D^{(\ell),R}\mathbf Q_t.
\end{equation}
This permits us to rotate the shared latent before caching while retaining layer-specific decoding.

\paragraph{Model-specific layouts and scoring.}
For DeepSeek-V3.2, we use one complex source per frequency, giving $r_R=d_R=64$, with native NeoX pairs $(0,32),\ldots,(31,63)$ and $\mathbf Q_t=\mathbf R_t$.
For GLM-5, we use two sources per frequency, giving $r_R=128$ and $d_R=64$. Each 64-dimensional source block uses native interleaved real/imaginary pairs, the original position angles, and rotary base $\theta=10^6$. In rank-major order,
\begin{equation}
\mathbf Q_t=\operatorname{diag}(\mathbf R_t,\mathbf R_t),
\qquad
\mathbf D^{(\ell),R}\in\mathbb R^{64\times128}.
\end{equation}

At inference, we absorb the decoders into post-RoPE, pre-Hadamard queries, matching the coordinate system of the reconstructed keys. We then quantize and score directly in latent coordinates, without applying Hadamard rotation or incorporating it into the decoders.

\subsection{Joint NoPE Regression}

For a group of $m$ layers and $n$ calibration tokens, let $\mathbf X\in\mathbb R^{n\times d}$ contain anchor inputs as rows, where $d=d_{\mathrm{model}}$, and concatenate the NoPE targets as
$\mathbf Y_N=[\mathbf K_{\ell_1,N},\ldots,\mathbf K_{\ell_m,N}]
\in\mathbb R^{n\times md_N}$.
We solve
\begin{equation}
\min_{\operatorname{rank}(\mathbf B_N)\le r_N,\,\mathbf b_N}
\frac1n\|\mathbf Y_N-\mathbf X\mathbf B_N-\mathbf1\mathbf b_N^\top\|_F^2
+\lambda_N\|\mathbf B_N\|_F^2.
\label{eq:ski_nope_objective}
\end{equation}
We penalize the effective mapping $\mathbf B_N$ and leave the bias unregularized, summing reconstruction errors across layers without energy normalization. Both models use $d_N=64$ and $r_N=128$.

Let $\boldsymbol\mu_x$ and $\boldsymbol\mu_y$ denote the training means, and define
\begin{equation}
\begin{aligned}
\mathbf C_{xx}
&=\mathbf X^\top\mathbf X/n-\boldsymbol\mu_x\boldsymbol\mu_x^\top,\\
\mathbf C_{xy,N}
&=\mathbf X^\top\mathbf Y_N/n-\boldsymbol\mu_x\boldsymbol\mu_y^\top,\\
\mathbf A_N&=\mathbf C_{xx}+\lambda_N\mathbf I.
\end{aligned}
\end{equation}
With $\mathbf V_N$ containing the leading $r_N$ right singular vectors of
$\mathbf A_N^{-1/2}\mathbf C_{xy,N}$, we obtain
\begin{equation}
\mathbf P_N=\mathbf A_N^{-1}\mathbf C_{xy,N}\mathbf V_N,
\qquad
\mathbf F_N=\mathbf V_N^\top,
\qquad
\mathbf B_N=\mathbf P_N\mathbf F_N,
\label{eq:ski_nope_solution}
\end{equation}
and
\begin{equation}
\mathbf b_N=\boldsymbol\mu_y-\mathbf B_N^\top\boldsymbol\mu_x.
\label{eq:ski_nope_bias}
\end{equation}
We split $\mathbf F_N$ into blocks $(\mathbf D^{(\ell),N})^\top$ and $\mathbf b_N$ into layer-specific biases. At inference, we project uncentered inputs as $\mathbf X\mathbf P_N$; the output biases incorporate the mean correction.

\subsection{Frequency-Wise Complex RoPE Regression}
\label{app:glm_rank2}

We pack each layer's pre-RoPE key pairs into complex coordinates. For frequency $f$, let $\mathbf Y_f\in\mathbb C^{n\times m}$ contain the group targets. We fit a bias-free, uncentered mapping:
\begin{equation}
\min_{\operatorname{rank}_{\mathbb C}(\mathbf B_f)\le r_f}
\frac1n\|\mathbf Y_f-\mathbf X\mathbf B_f\|_F^2
+\lambda_R\|\mathbf B_f\|_F^2.
\label{eq:app_rope_objective}
\end{equation}
Define
$\mathbf A_R=\mathbf X^\top\mathbf X/n+\lambda_R\mathbf I$
and
$\mathbf C_{xy,f}=\mathbf X^\top\mathbf Y_f/n$.
With $\mathbf V_f$ containing the leading $r_f$ right singular vectors of
$\mathbf A_R^{-1/2}\mathbf C_{xy,f}$, we obtain
\begin{equation}
\mathbf P_f=\mathbf A_R^{-1}\mathbf C_{xy,f}\mathbf V_f,
\qquad
\mathbf F_f=\mathbf V_f^H,
\qquad
\mathbf B_f=\mathbf P_f\mathbf F_f.
\label{eq:app_rope_solution}
\end{equation}
The columns of $\mathbf P_f$ generate the shared complex sources, and the corresponding layer column of $\mathbf F_f$ supplies the decoder coefficients $\gamma_{f,j}^{(\ell)}$. We fit frequencies independently, using $r_f=1$ for DeepSeek-V3.2 and $r_f=2$ for the final GLM-5 model.

\subsection{Regularization Selection and Numerical Implementation}
\label{app:ridge_selection}

\paragraph{Validation criterion.}
We search
$\mathcal A=\{0,10^{-8},10^{-7},10^{-6},10^{-5},10^{-4},10^{-3},10^{-2}\}$
and set
\begin{equation}
\lambda_c=\alpha_c\operatorname{tr}(\mathbf C_{xx})/d,
\qquad c\in\{N,R\}.
\end{equation}
Both branches use the anchor inputs' centered training variance to scale regularization, and all rotary frequencies within a group share one $\lambda_R$.

For each branch, we evaluate training-fitted parameters using the mean layer-wise normalized reconstruction error:
\begin{equation}
J_c(\alpha)=\frac1m\sum_{\ell\in\mathcal G_g}
\frac{\sum_{t\in\mathrm{val}}
\|\widehat{\mathbf k}_{\ell,t,c}(\alpha)-\mathbf k_{\ell,t,c}\|_2^2}
{\sum_{t\in\mathrm{val}}\|\mathbf k_{\ell,t,c}\|_2^2}.
\end{equation}
We accumulate errors and target energies over all validation tokens and retain the training-fitted NoPE bias. We select the strongest regularization within 0.5\% of the minimum validation error:
\begin{equation}
\alpha_c^\star
=\max\left\{\alpha\in\mathcal A:
J_c(\alpha)\le1.005\min_{\alpha'\in\mathcal A}J_c(\alpha')
\right\}.
\label{eq:ski_ridge_selection}
\end{equation}

\paragraph{Reported configurations.}
For DeepSeek-V3.2, we select regularization independently for each group and branch. All 15 groups in the 50-request calibration, $G=4$ configuration select $\alpha_N=\alpha_R=10^{-2}$. For $G=8$, we refit the groups using the same calibration and selection protocol.

For GLM-5, we first calibrate a configuration with one complex source per rotary frequency. We retain its NoPE parameters and refit the rotary projections and decoders with two sources per frequency, keeping each group's previously selected rotary regularization strength fixed.

\paragraph{Numerical implementation.}
We accumulate training and validation sufficient statistics separately in FP64, using token counts, feature sums, and the products $\mathbf X^\top\mathbf X$, $\mathbf X^\top\mathbf Y$, and $\mathbf Y^\top\mathbf Y$. We assemble complex rotary statistics from real-coordinate statistics and solve in complex128. This preserves all calibration observations without storing token-level features. We evaluate inverse operations spectrally, using a tolerance-based pseudoinverse at zero regularization and zero-padding rank-deficient factors to the prescribed dimensions. We check statistics-based validation errors against explicit predictions and verify exported FP32 parameters against the high-precision solution.

\section{Storage and Inference Implementation}
\label{app:storage_details}

\subsection{Cache Storage and Fixed Parameters}

Let $N_L$ denote the number of indexed layers, $N_{\mathrm{ind}}$ the number of independent native layers, $M$ the number of sharing groups, $L$ the history length, and $b$ the bytes per stored element. We compute cache payloads, excluding quantization metadata and allocation overhead, as
\begin{equation}
\mathcal M_{\mathrm{DSA}}=N_LLd_Ib,
\qquad
\mathcal M_{\mathrm{LatentIndex}}
=Lb(N_{\mathrm{ind}}d_I+Md_C).
\label{eq:ski_memory}
\end{equation}
For a group of $m$ layers, we replace $m$ native keys with one shared latent, giving a payload ratio of $d_C/(md_I)$.

In our DeepSeek-V3.2 FP8 implementation, we store each native key in 132 bytes: 128 FP8 values and one FP32 scale. Each shared latent occupies 200 bytes: 192 FP8 values and separate FP32 scales for the RoPE and NoPE segments. Including the independent native layer, we obtain
\begin{equation}
\begin{aligned}
m_{\mathrm{DSA}}&=61\times132=8052\ \text{bytes/token},\\
m_{G=4}&=132+15\times200=3132\ \text{bytes/token},\\
m_{G=8}&=132+8\times200=1732\ \text{bytes/token}.
\end{aligned}
\label{eq:ski_fp8_memory}
\end{equation}
We thus reduce logical indexer-cache storage by 61.1\% for $G=4$. This accounting excludes allocation overhead, temporary workspaces, HS candidate indices, fixed parameters, and main-attention KV caches.

We use $d_{\mathrm{model}}d_C$ shared projection parameters per group. Each group member additionally uses $d_Nr_N$ NoPE decoder parameters, $d_N$ bias parameters, and $2\sum_f r_f$ real rotary decoder parameters. These parameters are independent of context length.

\subsection{Query Absorption and FP8 Execution}

We generate the training-free latent through a linear projection without bias, activation, or additional normalization, using fitted targets that already include native key normalization. We apply RoPE to the rotary branch and cache the latent once per group. Each layer retains its native query and head-weight computations.

We perform decoder absorption on post-RoPE, pre-Hadamard queries. For NoPE, we compute
$(\mathbf D^{(\ell),N})^\top\mathbf q_N$
and the bias contribution
$\beta=\mathbf q_N^\top\mathbf b^{(\ell)}$.
For rotary frequency $f$ and latent source $j$, we compute
\begin{equation}
\widetilde q_{\ell,f,j}
=\overline{\gamma_{f,j}^{(\ell)}}q_{\ell,f}.
\end{equation}
We then score the rotated latent directly, retain the native indexer scaling, and add $\beta$ before ReLU and head aggregation. We neither apply Hadamard rotation on this path nor incorporate it into the decoders.

For DeepSeek-V3.2, we use a RoPE64/NoPE128 layout. We quantize cached latents and absorbed queries separately by segment using E4M3FN, with independent scales rounded up to powers of two; cached key scales are stored in FP32. The kernel rescales the segment dot products, sums them, adds the bias contribution, applies ReLU, and aggregates heads. We export calibration parameters in FP32, perform online latent projection in BF16, and compute query absorption in FP32 before conversion to BF16 and FP8 quantization.

For GLM-5, we use a RoPE128/NoPE128 segmented FP8 layout. The rotary segment comprises two consecutive 64-dimensional source blocks with native interleaved real/imaginary pairs. We apply the same native cosine and sine values to both blocks and quantize them together as one rotary segment.

\subsection{Computation With and Without Hierarchical Selection}

Without HS, we independently score the full causal history and select top-$k$ tokens at every layer. NoPE query absorption costs $O(H_Id_Nr_N)$ per query, and rotary absorption costs $O(H_Ir_R)$. Scoring costs $O(H_ILd_C)$ per layer per query and $O(H_IL^2d_C)$ over prefill. Each layer separately reads the shared cache and scores in the $d_C$-dimensional latent space.

With HS, we let the anchor score the full prefix and each follower score at most $B$ candidates. For a group of $m$ layers, we therefore score
\begin{equation}
L+(m-1)\min(B,L)
\end{equation}
positions per query instead of $mL$. For DeepSeek-V3.2 with one independent layer, 15 anchors, 45 followers, and $B=8192$, this gives $16L+45\times8192$ positions when $L>8192$, compared with $61L$ for native DSA.

These counts cover scoring positions only. In Appendix~\ref{app:hs_timing}, we measure complete indexer forward calls, including candidate construction and TopK, while excluding main attention and other model components.

\section{Training-Free Experimental Details}
\label{app:experimental_details}

\subsection{Models and Comparison Settings}

We compare native DSA, IndexCache, and LatentIndex on DeepSeek-V3.2 and GLM-5. Native DSA selects tokens independently at each indexed layer. For IndexCache, we reuse anchor selections with the same group boundaries as LatentIndex. All methods select $K=\min(2048,N)$ tokens for a valid causal prefix of length $N$.

\begin{table}[t]
\centering
\caption{Training-free model and latent configurations. Layer indices are zero-based.}
\label{tab:model_config}
\small
\begin{tabular}{lcc}
\toprule
Setting & DeepSeek-V3.2 & GLM-5 \\
\midrule
Indexed layers $N_L$ & 61 & 78 \\
Hidden dimension $d_{\mathrm{model}}$ & 7168 & 6144 \\
Indexer heads $H_I$ & 64 & 32 \\
Native RoPE / NoPE $(d_R,d_N)$ & 64 / 64 & 64 / 64 \\
Shared RoPE / NoPE $(r_R,r_N)$ & 64 / 128 & 128 / 128 \\
Independent layer indices & 0 & 0, 1 \\
Grouped layer indices & 1--60 & 2--77 \\
Number of four-layer groups & 15 & 19 \\
Complex sources per rotary frequency & 1 & 2 \\
\bottomrule
\end{tabular}
\end{table}

We use $G=4$, 50-request calibration, and segmented FP8 inference for the main results (Table~\ref{tab:model_config}). We keep fitted parameters fixed throughout evaluation and use the GLM-5 calibration procedure described in Appendix~\ref{app:ridge_selection}.

\subsection{Downstream Evaluation and Recall Protocol}

\paragraph{Downstream evaluation.}
We evaluate RULER on all 13 tasks with 100 examples per task at each of six context lengths: 4K, 8K, 16K, 32K, 64K, and 128K. We average task scores within each length and then equally across lengths. For LongBench, we evaluate six categories: single-document QA, multi-document QA, summarization, few-shot learning, synthetic retrieval, and code completion. We use completion mode for code tasks and chat mode otherwise, averaging task scores within each category and then equally across categories. We run each method end to end using its own token selections.

\paragraph{Recall sampling and aggregation.}
At each nominal context length, we use 100 examples from each of the 13 RULER tasks. We evaluate the query at the first token of the question text, using only the causal prefix through that position; $N$ denotes its actual length. For each example, we run the native DSA trajectory and compare all selectors at the same query using the same dense core-attention probabilities, computed over the full causal prefix with the native attention scale.

We compute head-wise oracle-normalized ratios as defined in Section~\ref{sec:selection_quality}, average over core-attention heads, and then over all indexed layers, including independent layers: 61 for DeepSeek-V3.2 and 78 for GLM-5. We subsequently average examples within each task and give tasks equal weight. Length averages give the six lengths equal weight. For DeepSeek-V3.2, we use all 128 core-attention heads.

\subsection{Cross-Layer Sharing Analysis}
\label{app:additional_analysis}

\paragraph{Group offsets and group size.}
For the DeepSeek-V3.2 $G=4$ analysis, we measure additional recall loss relative to each layer's native DSA selection on the same reference trajectory. We average layers at the same offset across the 15 groups, with offset 0 denoting anchors and offsets 1--3 denoting followers, then aggregate examples and tasks as above. We include independent layer 0 in overall recall but exclude it from offset analysis.

For $G=8$, we retain layer 0 independently, form seven eight-layer groups from layers 1--56, and place layers 57--60 in one four-layer group. We retain the RoPE64/NoPE128 latent layout and refit projections and decoders using the same calibration protocol. We use identical group boundaries for IndexCache.

\begin{figure}[t]
\centering
\includegraphics[width=0.5\linewidth]{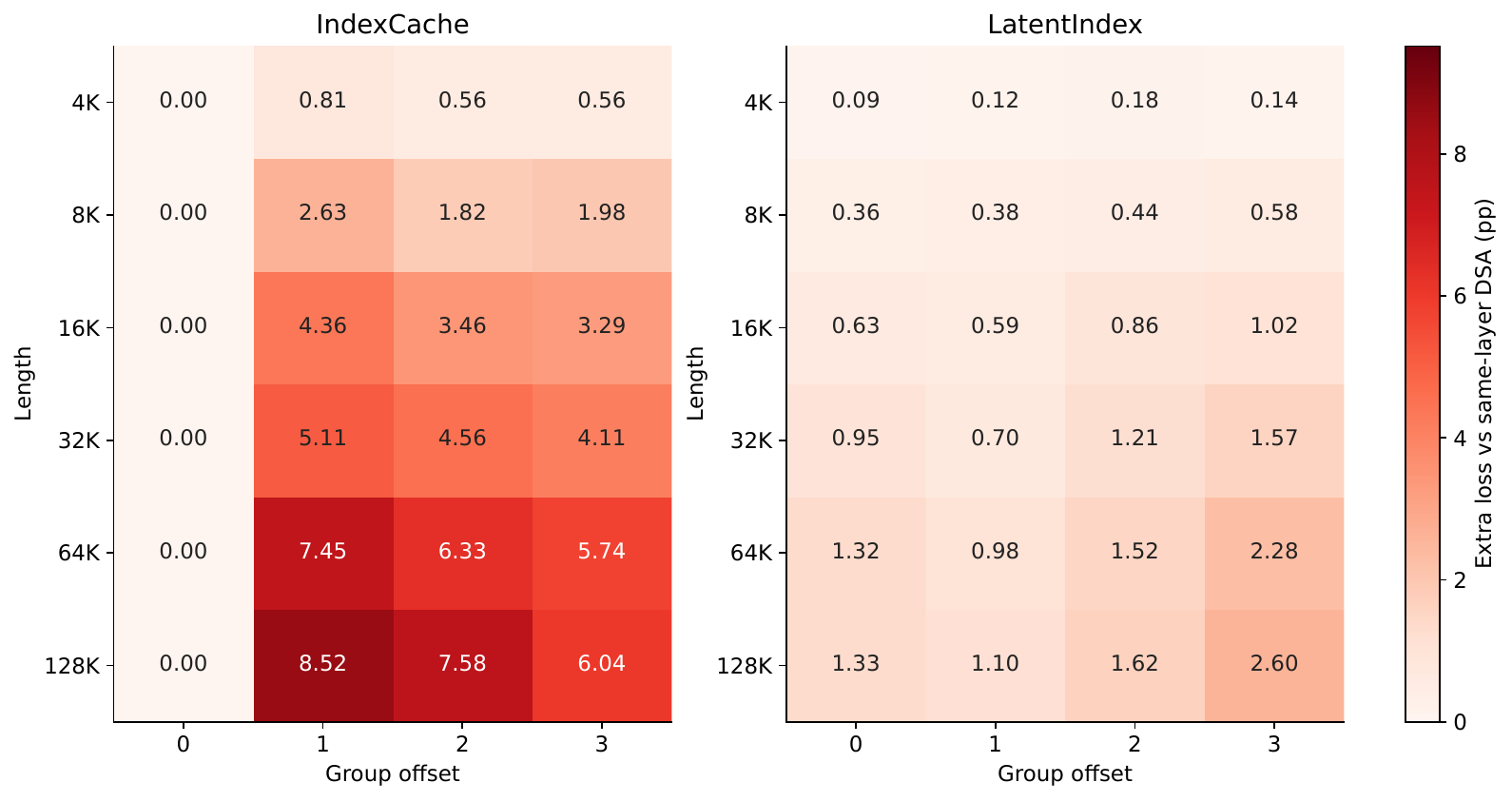}
\caption{Additional $R_{\mathrm{mean}}$ loss relative to native DSA by group offset on DeepSeek-V3.2 ($G=4$). We use the same trajectory, sampling, and aggregation as in Figure~\ref{fig:recall_loss}.}
\label{fig:rmean_appendix}
\end{figure}

\paragraph{Mean-attention recall.}
Using the same dense probabilities as $R_{\mathrm{head}}$, we first average probabilities across heads as $\bar p_i=H^{-1}\sum_h p_{h,i}$ and define the oracle $T=\operatorname{TopK}_i(\bar p_i,K)$. We compute
\begin{equation}
R_{\mathrm{mean}}
=
\frac{\sum_{i\in S}\bar p_i}
{\max\!\left(\sum_{i\in T}\bar p_i,\epsilon\right)},
\qquad \epsilon=10^{-8}.
\end{equation}
This metric measures how well a token set shared across attention heads preserves head-averaged attention mass relative to the optimal shared size-$K$ set. We use it to assess whether the advantage observed under head-wise normalization also holds for aggregate attention mass. We compute both metrics in the same evaluation pass and apply the same layer, example, and task aggregation. Figure~\ref{fig:rmean_appendix} reports additional $R_{\mathrm{mean}}$ loss relative to native DSA using the same offset analysis.

\subsection{Hierarchical Selection and Decode Indexer Timing}
\label{app:hs_timing}

\paragraph{Candidate construction.}
We use $B=8192$ candidates and final selection budget $k=2048$ at all lengths. Layer 0 retains native DSA. In each four-layer group, the anchor scores the full prefix and constructs its candidate set from its top-2048 positions and the next 6144 highest-scoring positions. Each follower independently rescores the candidates and selects its own top-2048. We cap both budgets by the valid prefix length and rebuild candidates for each group at every decode step. At 4K, followers score the full valid prefix; at nominal 8K, measured decode prefixes exceed 8192 tokens and use the fixed candidate budget.

\paragraph{Workload and aggregation.}
We measure batch-size-one decoding on eight GPUs with tensor parallelism 8, using nested prefixes of the same book. Each request generates 65 tokens. We exclude the first token produced by prefill and measure the following 64 decode steps, whose history lengths are approximately $L+1$ through $L+64$ for nominal input length $L$. We measure three requests per method and length.

We place CUDA events around each layer's complete \texttt{indexer.forward}. Let $t_{\ell,r}^{(j,s)}$ denote elapsed time for layer $\ell$, TP rank $r$, request $j$, and decode step $s$. We report
\begin{equation}
\overline T_{\mathrm{indexer}}
=\frac13\sum_{j=1}^{3}\frac1{64}\sum_{s=1}^{64}
\sum_{\ell=0}^{60}\max_{r=0,\ldots,7}t_{\ell,r}^{(j,s)}
\label{eq:app_indexer_timing}
\end{equation}
in milliseconds per decode token.

\section{Training-Aware Experimental Details}
\label{app:training_aware_details}

\subsection{Training Configuration}
\label{app:training_details}

\begin{table}[t]
\centering
\caption{Training settings. All methods completed both schedules. Batch sizes count sequences; the global indexer-loss coefficient is distinct from the method-specific layer weights in Appendix~\ref{app:training_implementation}.}
\label{tab:training_aware_settings}
\small
\begin{tabular}{lcc}
\toprule
Setting & Dense warmup & Sparse training \\
\midrule
Sequence length & 65,536 & 65,536 \\
Completed optimizer steps & 400 & 2,426 \\
Global batch size & 8 & 256 \\
Microbatch size & 1 & 1 \\
Gradient accumulation steps & 8 & 4 \\
Learning rate & $10^{-3}$ & $4\times10^{-5}$ \\
Learning-rate schedule & Constant & Constant \\
Learning-rate warmup steps & 0 & 0 \\
Indexer objective & Full-prefix KL & Selected-support KL \\
Global indexer-loss coefficient & 1.0 & 1.0 \\
Backbone updates & Disabled & Enabled \\
Precision & BF16 & BF16 \\
GPU count & 8 & 64 \\
TP / PP / CP / EP & 1 / 1 / 8 / 8 & 1 / 1 / 1 / 8 \\
\bottomrule
\end{tabular}
\end{table}

We compare DSA, IndexCache, and LatentIndex on DeepSeek-V2-Lite using eight indexer heads and top-2048 selection. For grouped methods, we partition the 27 layers into nine groups of three. 
Dense warmup freezes the backbone and trains only the indexers. Sparse training loads each method's own warmup checkpoint, activates its routes, and updates the backbone with next-token LM loss while continuing indexer supervision.
We initialize new indexer linear weights from a zero-mean normal distribution with standard deviation 0.02, and LayerNorm weights and biases to one and zero, respectively. 
For sparse training, we use the same dataset, cached sample and shuffle indices, seed, and sequential sampler across methods. We use a global batch size of 256, microbatch size 1, data parallelism 64, and context parallelism 1, aligning sample slots at matched training steps.

\subsection{Indexer Construction and Training Objectives}
\label{app:training_implementation}

\paragraph{Shared latent indexers.}
For each group, we project the anchor's normalized hidden state into a 128-dimensional NoPE latent and a 64-dimensional RoPE key. We apply LayerNorm only to the NoPE latent and use layer-specific linear decoders to produce 64-dimensional NoPE keys. Each layer directly reuses the shared RoPE key, corresponding to an identity rotary decoder.

We generate all three layers' queries and head weights from the anchor hidden state using separate layer-specific projections. All linear projections are bias-free. After applying RoPE, we apply the same normalized Hadamard rotation to queries and keys. During sparse training, we compute three independent top-2048 selections at the group entrance for the corresponding layers to consume. During warmup, we evaluate each layer's indexer output against its own dense-attention teacher using the cached anchor input.

\paragraph{Attention supervision.}
For layer $\ell$ and query $t$, we form the teacher distribution $P_t^{(\ell)}$ by averaging the 16 core-attention heads' probabilities, using the runtime attention scale, including YaRN scaling. We form the indexer distribution $S_t^{(\ell)}$ by applying softmax to its scores. The indexer's head weights include the scaling factor $1/(\sqrt{8}\sqrt{128})$.

During warmup, both distributions cover the full causal prefix. During sparse training, we use the current layer's sparse-attention probabilities as the teacher and normalize both distributions over the selected positions. With prediction mask $m_t$, we define
\begin{equation}
\overline{\mathcal L}_{\ell}
=
\frac{\sum_t m_t
D_{\mathrm{KL}}\!\left(
\operatorname{sg}[P_t^{(\ell)}]
\,\|\,S_t^{(\ell)}
\right)}
{\sum_t m_t},
\end{equation}
where $\operatorname{sg}$ denotes stop-gradient. We detach indexer inputs from the backbone and optimize the backbone through next-token language modeling during sparse training.

\paragraph{Layer-wise loss normalization.}
All methods receive supervision from all 27 layers. DSA and LatentIndex use each layer's corresponding indexer output, while IndexCache supervises the shared anchor indexer separately with each group member's teacher. In both stages, we use
\begin{equation}
\mathcal L_{\mathrm{idx}}
=\sum_{\ell=0}^{26}c_\ell\overline{\mathcal L}_{\ell},
\qquad
c_\ell=
\begin{cases}
1/3, & \text{IndexCache},\\
1, & \text{DSA and LatentIndex}.
\end{cases}
\end{equation}
We apply the training framework's microbatch and data-parallel reductions without an additional division by the number of layers or groups.

\subsection{Recall Diagnostics and Plotted Training Curves}
\label{app:training_diagnostics}

\paragraph{Warmup Recall@2048.}
For each sampled query $s$ and layer $\ell$, we measure selected attention mass relative to each head's top-$K$ oracle, where $K=\min(2048,N)$ for causal prefix length $N$:
\begin{equation}
r_{\ell,s}
=\frac1{16}\sum_{h=1}^{16}
\frac{\sum_{t\in\mathcal I_{\ell,s}}
\widetilde p_{s,t,h}^{(\ell)}}
{\sum_{t\in\mathcal O_{\ell,s,h}}
\widetilde p_{s,t,h}^{(\ell)}}.
\end{equation}
Here $\mathcal I_{\ell,s}$ is the diagnostic indexer selection, $\mathcal O_{\ell,s,h}$ is the head-specific oracle, and $\widetilde p$ is the diagnostic dense-attention distribution.

We sample at most two evenly spaced local queries per layer and context-parallel rank, using CP=8. We average query-level recall within each layer, then across ranks and all 27 layers. All methods use the same sampling procedure and data order. For IndexCache followers, we score with the anchor's queries, keys, and head weights against each follower's own dense-attention reference.

We compute diagnostic attention probabilities with scale $1/\sqrt{576}$, corresponding to the absorbed MLA query width $512+64$. The diagnostic therefore uses the head-wise oracle normalization of $R_{\mathrm{head}}$, with its own attention scaling and query sampling.

\paragraph{LM loss and plotting.}
We report next-token cross-entropy over valid prediction tokens, with EOD loss masking and excluding auxiliary losses. We normalize within each microbatch by its valid-token count, then average across microbatches and data-parallel ranks. Each logged value corresponds to its optimizer step. We compute method-minus-DSA differences at matched steps without interpolation and apply bias-corrected causal EMA after differencing, using coefficients 0.9 for warmup and 0.6 for sparse training.

\end{document}

%% file: math_commands.tex
\usepackage{amsmath,amsfonts,bm}

\def\eqref#1{equation~\ref{#1}}

\def\1{\bm{1}}

\DeclareMathAlphabet{\mathsfit}{\encodingdefault}{\sfdefault}{m}{sl}
\SetMathAlphabet{\mathsfit}{bold}{\encodingdefault}{\sfdefault}{bx}{n}

